\documentclass[letterpaper, 10pt]{article}
\usepackage[a4paper,top=2cm,bottom=2cm,left=2.5cm,right=2.5cm,marginparwidth=1.75cm]{geometry}

\makeatletter
\DeclareRobustCommand{\ch}{%
  \@bsphack
  \leavevmode 
  \color{red} 
  \@esphack
}
\DeclareRobustCommand{\stopch}{%
  \@bsphack
  \normalcolor
  \@esphack
}
\makeatother

\usepackage{times}
\usepackage{epsfig}
\usepackage{graphicx}
\usepackage{amsmath}
\usepackage{amssymb}
\usepackage{booktabs}
\usepackage{textcomp}
\usepackage{subcaption}
\usepackage{soul}
\usepackage{algorithm}
\usepackage{algpseudocode}
\usepackage{paralist}

\usepackage[inline]{enumitem}
\usepackage[table]{xcolor}
\usepackage{multirow}
\usepackage{tabularx}
\usepackage{tablefootnote}
\usepackage{url}
\usepackage{makecell}
\usepackage[colorlinks]{hyperref}

\newcommand{\eg}{\textit{e.g.}, }
\newcommand{\etal}{\textit{et al.}}

\newcommand{\nostarnote}[1]{}

\title{\LARGE \bf The Common Objects Underwater (COU) Dataset for Robust Underwater Object Detection}
\author{Rishi Mukherjee$^{1}$, Sakshi Singh$^{2}$, Jack McWilliams$^{3}$, Junaed Sattar$^{4}$
\thanks{This work was supported in part by the United States National Science Foundation grant \#IIS-2220956.}%
\thanks{Authors are with the Department of Computer Science \& Engineering and the Minnesota Robotics Institute (MnRI),
        University of Minnesota--Twin Cities, Minneapolis, MN 55455, USA {\tt\small \{$^{1}$mukhe100,$^{2}$sing0975, $^{3}$mcwil085, $^{4}$junaed\}@umn.edu}}}%
\date{}
\begin{document}

\maketitle
\thispagestyle{empty}
\pagestyle{empty}

\begin{abstract}
    We introduce COU: Common Objects Underwater, an instance-segmented image dataset of commonly found man-made objects in multiple aquatic and marine environments. 
    COU contains approximately $10$K segmented images, annotated from images collected during a number of underwater robot field trials in diverse locations. 
    COU has been created to address the lack of datasets with robust class coverage curated for underwater instance segmentation, which is particularly useful for training light-weight, real-time capable detectors for Autonomous Underwater Vehicles (AUVs). 
    In addition, COU addresses the lack of diversity in object classes since the commonly available underwater image datasets focus only on marine life.
    Currently, COU contains images from both closed-water (pool) and open-water (lakes and oceans) environments, of $24$ different classes of objects including marine debris, dive tools, and AUVs.
    To assess the efficacy of COU in training underwater object detectors, we use three state-of-the-art models to evaluate its performance and accuracy, using a combination of standard accuracy and efficiency metrics. 
    The improved performance of COU-trained detectors over those solely trained on terrestrial data demonstrates the clear advantage of training with annotated underwater images. 
    We make COU available for broad use under open-source licenses. 
\end{abstract}

\section{Introduction}
\label{sec:introduction}

There is an increasing demand for robust underwater object detection systems, particularly for autonomous underwater vehicles (AUVs), that can benefit various fields such as marine conservation, search and rescue, and environmental monitoring \cite{industryarc2025, searchandrescue, auvsearchandrescue, marine_conservation}. 
A dataset curated specifically for underwater object detection can be a valuable resource for training and testing models intended for these applications. 
Particularly for underwater robotics, the need for training robust and efficient object detectors is paramount; this can improve the accuracy of underwater navigation, marine biodiversity tracking, environmental change detection, and the monitoring of human impact on underwater ecosystems.

While existing datasets like COCO (Common Objects in Context) \cite{COCO} have significantly advanced object detection for terrestrial scenes, they have inherent limitations when applied to underwater contexts, leading to reduced accuracy in detecting and segmenting underwater objects. 
This phenomenon is observed in our own evaluations of object detectors trained solely on terrestrial object datasets (see Sec.~\ref{sec:results}).
In this paper, we introduce the COU (Common Objects Underwater) dataset, the first of its kind consisting solely of underwater object images from pools, lakes, and ocean environments.
COU contains segmented and bounding-box annotated images of $24$ classes of objects (Sec.~\ref{sec:creation}) including objects commonly used by divers, and marine debris.
These objects represent man-made objects which commonly end up as debris, derelict fishing gear, dive equipment, and even underwater robotic vehicles. 
COU stems from the need to address the growing need for high-quality, domain-specific data to improve the performance of object detection models in a variety of underwater environments, and provides several advantages for underwater detector training:
\begin{compactenum} 
\item Most common object detection datasets containing large amount of images (\eg COCO, ImageNet \cite{imagenet}) focus on objects in everyday terrestrial scenes, which vary significantly from underwater environments.
Underwater imaging presents unique challenges such as color distortion, light refraction, limited visibility, and suspended particles~\cite{lu2017underwater}, which are absent or less prevalent in typical terrestrial object detection datasets.
COU addresses these environmental variables by ensuring that all images are captured underwater, creating a more relevant and reliable dataset for training models specifically tailored to these conditions.  
\item Models trained on terrestrial datasets often suffer from domain shift when applied to underwater environments, where object appearance, background, and lighting differ dramatically. 
COU mitigates this domain shift by focusing solely on underwater scenes, allowing object detection models to better generalize to underwater conditions, leading to improved performance for applications such as marine biology, ocean exploration, and underwater human-robot interaction (UHRI). 
\item Terrestrial object categories are limited in scope and do not encompass the wide variety of objects commonly found underwater, such as underwater structures or human-made debris. 
COU expands on this by annotating a wide range of underwater objects (\eg pollution: plastic bags, bottles, ROVs, dive equipment: snorkels, flippers, goggles, dive weights). 
This ensures that models trained on COU can recognize a diverse set of underwater objects, filling the gap left by more general categories in terrestrial datasets. 
\item The clarity and definition of object boundaries in underwater images are often compromised due to the scattering of light and reduced visibility. 
COU focuses on precise segmentation of objects in underwater scenes, providing high-quality, pixel-level annotations that account for the unique visual characteristics of underwater objects. 
This refinement allows for more accurate object segmentation, which would particularly assist in tasks requiring fine-grained analysis, such as species identification or object retrieval. 
\end{compactenum}

\begin{figure}
    \centering
    \subfloat[]{
        \includegraphics[width=0.49\linewidth]{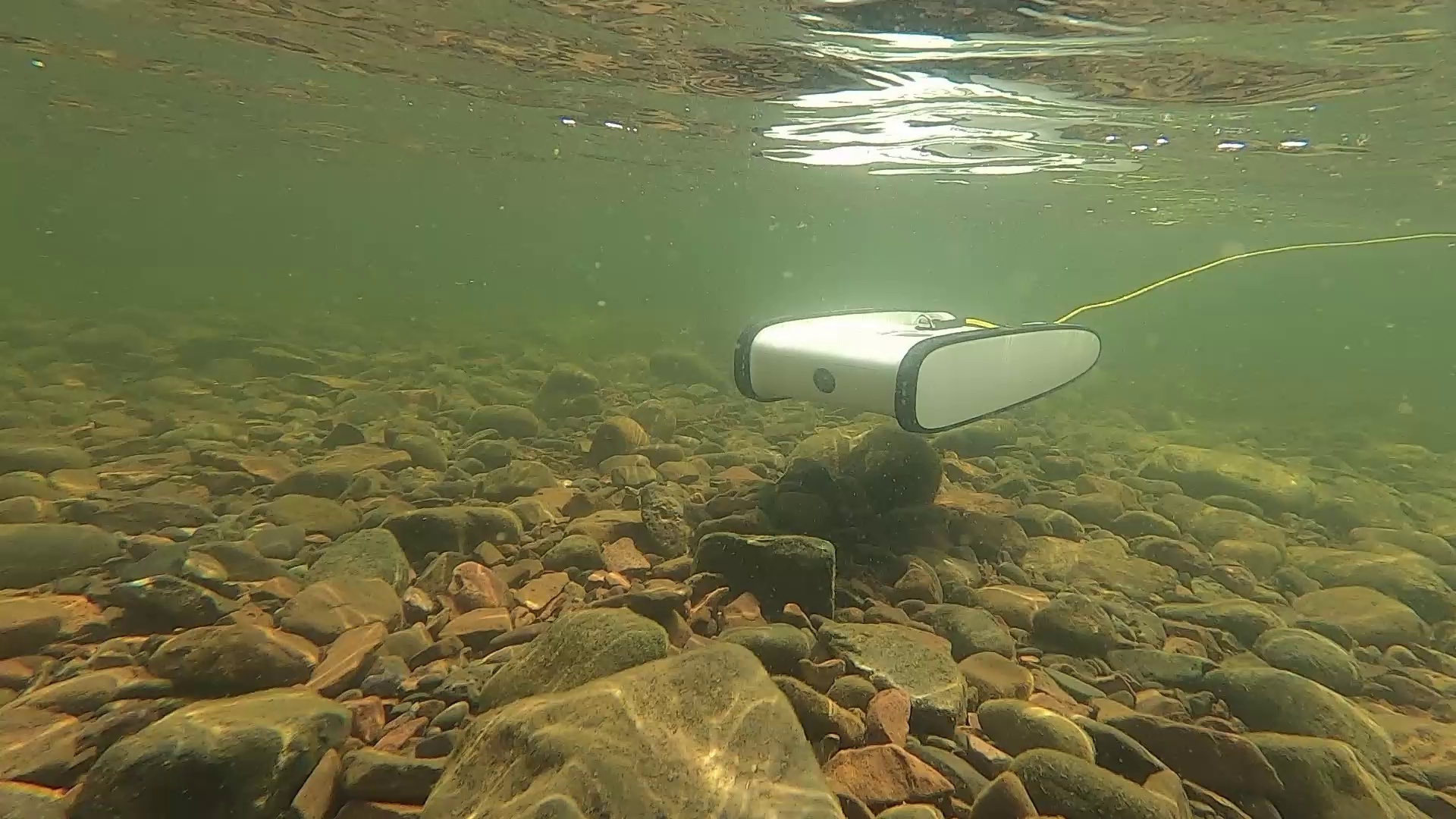}
        \label{fig:intro_image1}
    }    
    \subfloat[]{
        \includegraphics[width=0.49\linewidth]{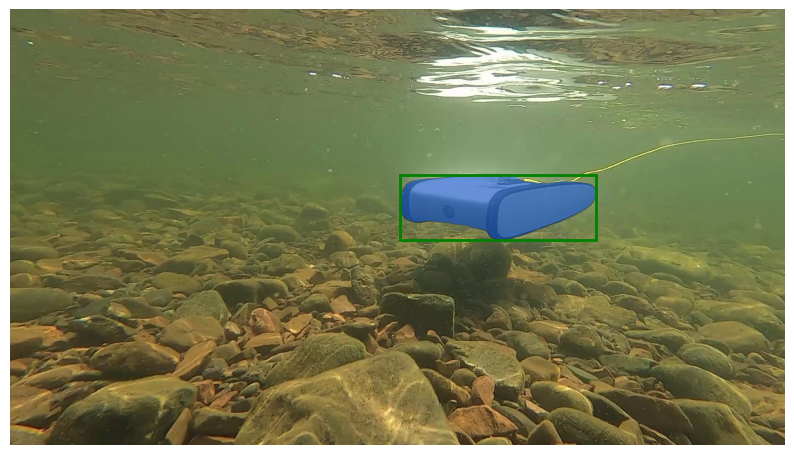}
        \label{fig:intro_image2}
    }
    \caption{Sample images from COU, captured in Lake Superior, USA. (a) The raw image is of a Trident AUV, with (b) COU containing the segmentation mask and bounding box of the ``ROV'' class.}
    \label{fig:intro_image}
\end{figure}

Consequently, our efforts in curating the COU dataset have prioritized these factors. 
Images from various camera sources, at good resolution, and from diverse underwater environments (geographically ranging from Lake Superior to the Caribbean Sea) have been used in COU to present $24$ classes of objects (Fig. \ref{fig:intro_image}). 
While not particularly large compared to terrestrial equivalents, COU still provides sufficient diversity in annotated underwater objects, a resource the underwater vision and robotics communities currently lack. The data is available at: https://z.umn.edu/cou-dataset

\section{Related Work}

Over the years, deep learning models have played a crucial role in the advancement of object detection and recognition and culminated in some of the most accurate and robust are object detectors to date~\cite{wang2024yolov9learningwantlearn, carion2020endtoendobjectdetectiontransformers, he2018maskrcnn}.
The majority of the datasets are trained on large terrestrial object datasets like COCO \cite{COCO}, ImageNet \cite{imagenet}, and PASCAL VOC \cite{pascal_voc}. 
COCO contains $330$K images and $1.5$ million object instances. 
ImageNet contains approximately $14$ million images along with their bounding box annotations. 
PASCAL VOC contains approximately $11$k images. 
The variety and vastness of data available for terrestrial applications have played a pivotal role in the advancement of the field.

Unfortunately, a model trained on terrestrial data can not be transferred to the underwater domain directly and at least needs to be fine-tuned on some underwater data. 
This is due to the changes in visibility, light diffraction, and erosion of objects underwater. 
Image enhancement methods \cite{funie, enhance_densegan} have been shown to improve the detection performance by accounting for some of the challenges underwater. 
While these methods aid in the model performance, they do not change the image enough to resemble terrestrial data, necessitating the use of actual underwater data for training. 
To account for the underwater appearance, \cite{submergegan} uses generative techniques to augment terrestrial data. 
While the performance in this case is better than training the model on only terrestrial data, it still lags when compared to the performance of these models on datasets like COCO. 
This further emphasizes the need for good training data.

There are some underwater datasets available for training detectors; \eg VDD-C~\cite{vddc} contains images of divers with their bounding box annotations and contains $105$k annotated images of divers and swimmers in both pool and ocean environments. 
While it is large enough, it has a limited number of classes and does not cover any objects and organisms present in the ocean. 
CADDY~\cite{caddy} also contains diver-focused data. 
Additionally, the Brackish~\cite{pedersen2019brackish} dataset contains $14,674$ images out of which $12,444$ contain objects with bounding box annotations. 
It has $5$ classes including fish, crabs, and other marine life, collected at a depth of $9$ meters.
The Moorea Labeled Corals dataset~\cite{coralnet} contains $400,000$ annotations.
The DUO (Detecting Underwater Objects) dataset~\cite{duo} contains images from $2$ datasets. 
It contains over $8$k images of marine life and other small organisms from the datasets from Underwater Robot Picking Competition (URPC) and UDD \cite{udd}.
DUO augments and combines two datasets and provides better annotations and training data, and contains $4$ classes of marine animals.
Additionally, RoboFlow \cite{roboflow_aq} contains $638$ images containing $7$ classes of aquatic animals with bounding box annotations. 
This dataset contains images from two aquariums.
While these datasets contain a vast amount of images from marine life and coral, they do not contain annotations or images of debris or man-made objects underwater.

OUC-Vision~\cite{ouc_vision} is an underwater image database for saliency detection, which contains $4,400$ underwater images of $220$ individual objects, each captured with $4$ pose variations and $5$ spatial locations, under varying lighting conditions. 
Fulton et.al~\cite{trash19} extracts $5,700$ images from video sequences and labels them with bounding box annotations. 
It includes instances of trash, marine animals and plants, and remote-operated vehicles (ROVs).
This dataset was further expanded to form TrashCan\cite{trashcan}, which contains $7,212$ images of trash, marine life, and ROVs. 
It includes pixel-level annotations along with bounding box information. 
UNO \cite{UNO} further fine-tunes the annotations in the TrashCan dataset and contains $5,902$ image frames and $10,773$ annotations. 
Seaclear marine debris dataset \cite{tudelft} contains $8,610$ images across three: debris, marine life, and robot. 
Additionally, Zocco \etal ~\cite{data_2class} introduces a plastic waste dataset containing $900$ images with two classes, plastic bags and plastic bottles. 
DeepPlastic \cite{tata2021deepplastic} contains $3,174$ images of marine-based plastic.

Most of these datasets, however, either cover marine life or divers. 
Those that cover non-organic objects and trash present underwater are limited in the number of images and the environments where data is captured. 
COU aims to tackle both issues. 
We use images of objects in pools, lakes, and ocean environments. 
We also aim to include $24$ classes of objects present with divers (\eg goggles, flippers) and trash that are often present underwater. 
We provide both bounding box annotations and segmented annotations of the objects, expanding our dataset to currently contain $9,757$ images.

\section{Dataset Creation and Processing}
\label{sec:creation}
One key motif for creating COU is to maximize the diversity of lighting conditions and environments, to enable robust use onboard underwater robots. 
Different aquatic environments can have drastically different hues, while poor lighting conditions can cause the edges of objects to merge with the background. 
The data collection and processing tasks, accordingly, prioritized these requirements, as described in Section \ref{env_select}.

\subsection{Object Selection}
\begin{figure}[t]
    \centering
    \includegraphics[width=0.9\textwidth]{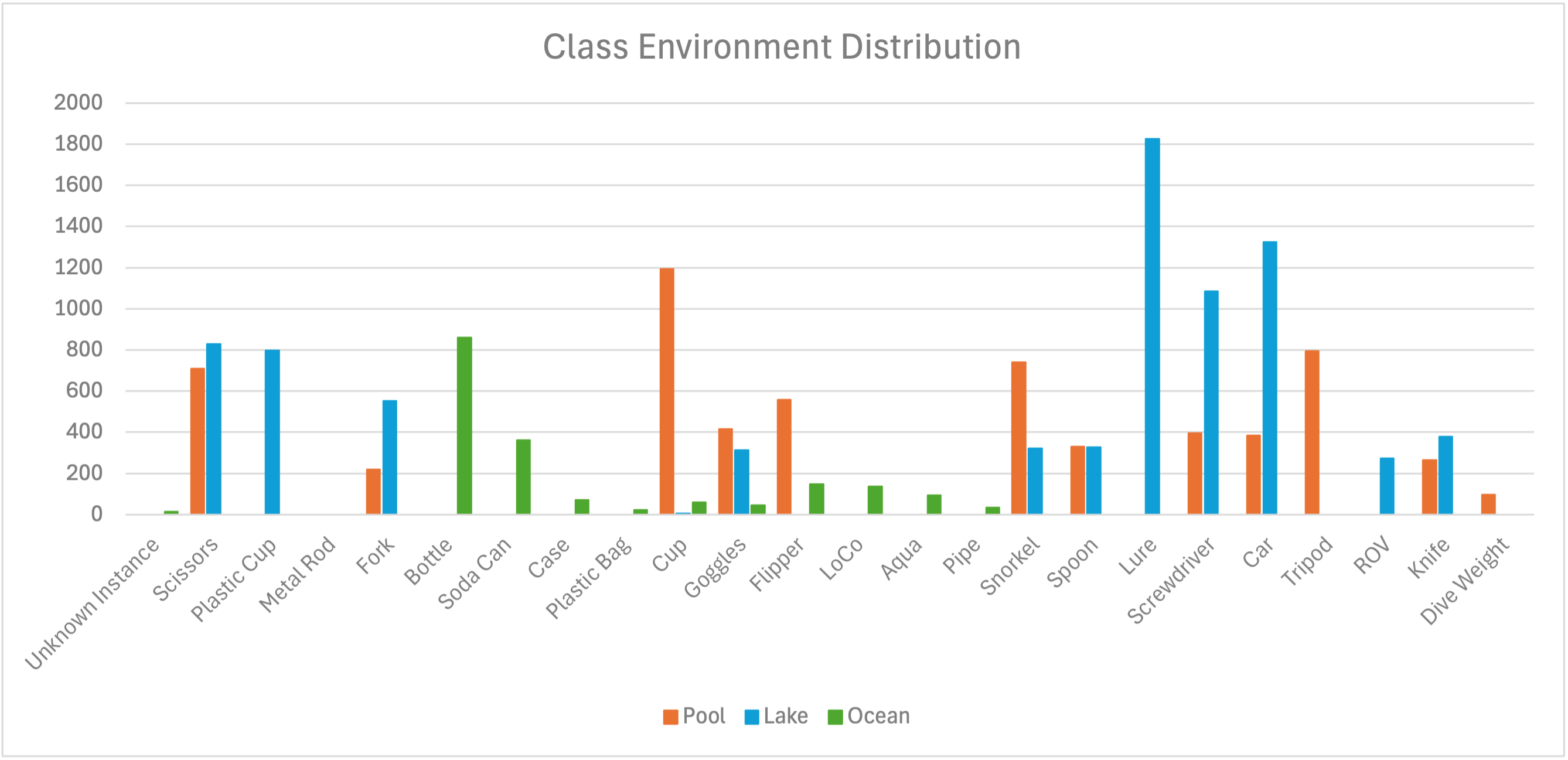}
    \caption{Distribution of object classes showing their respective environments. \emph{orange: Pool, blue: Lake, green: Ocean}}
    \label{fig:class-env-dist}
\end{figure}

The objects chosen in COU represent two key objectives: common objects present in the underwater domain, and objects that are relevant for human-robot collaboration. We use COCO \cite{COCO} as a guide for selecting common objects - such as scissors, screwdrivers, etc. to be filmed in the underwater domain, and for diver-AUV collaboration, we add instances of swim gear - such as goggles, flippers, snorkel, etc. as well as some publicly available AUV platforms in the dataset. We do not classify marine bio-life or marine debris as there are existing datasets that cater to those kinds of classes \cite{trash19} \cite{coralnet}. Fig ~\ref{fig:class-env-dist} contains the full list of objects and their environments for building the dataset.

\subsection{Environment Selection}
\label{env_select}
\begin{figure}[htbp]

    \centering
    \begin{subfigure}[b]{0.33\linewidth}
        \centering
        \includegraphics[width=\textwidth]{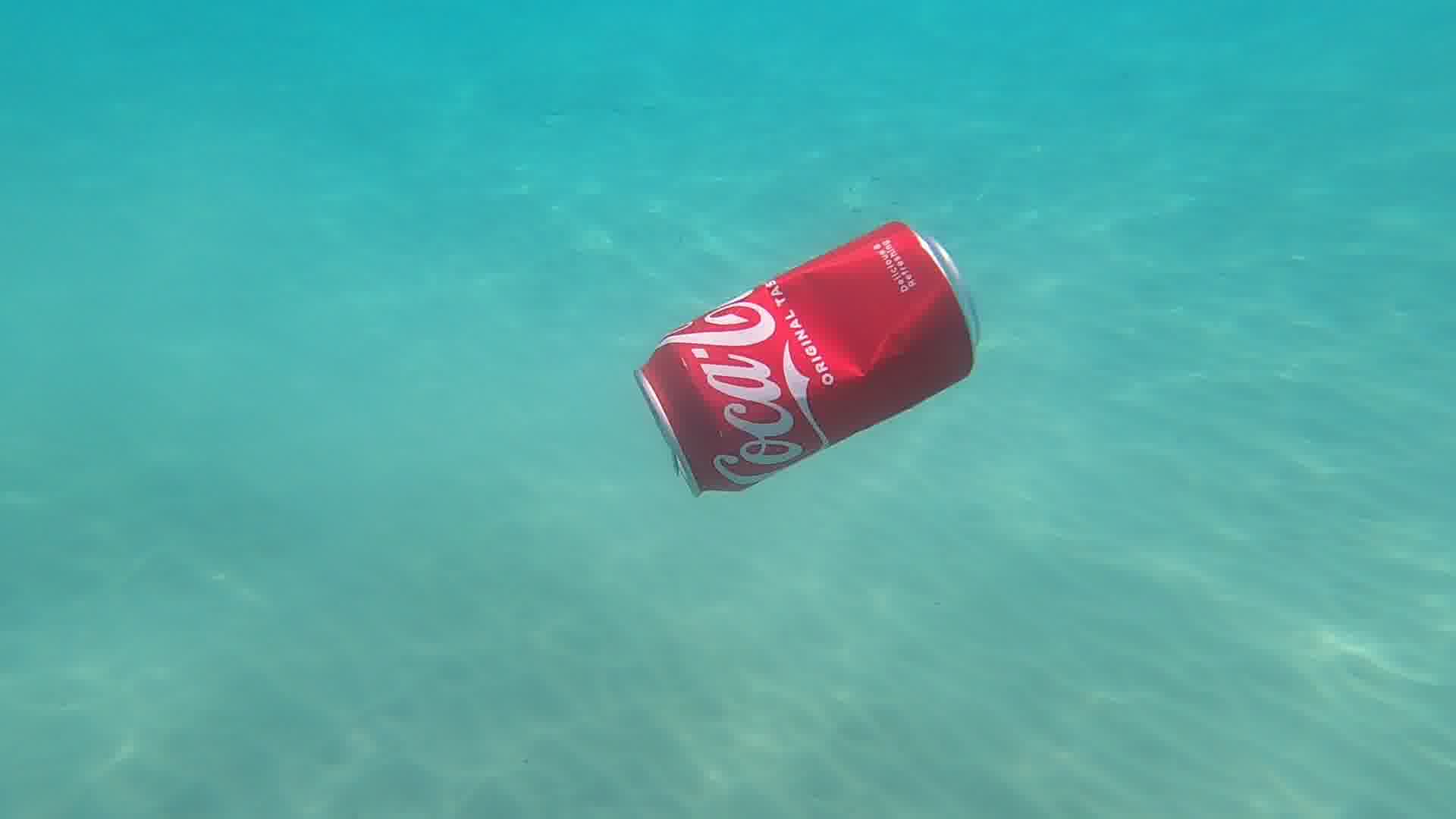}
        \caption{Ocean}
        \label{fig:image11}
    \end{subfigure}%
    \begin{subfigure}[b]{0.33\linewidth}
        \centering
        \includegraphics[width=\textwidth]{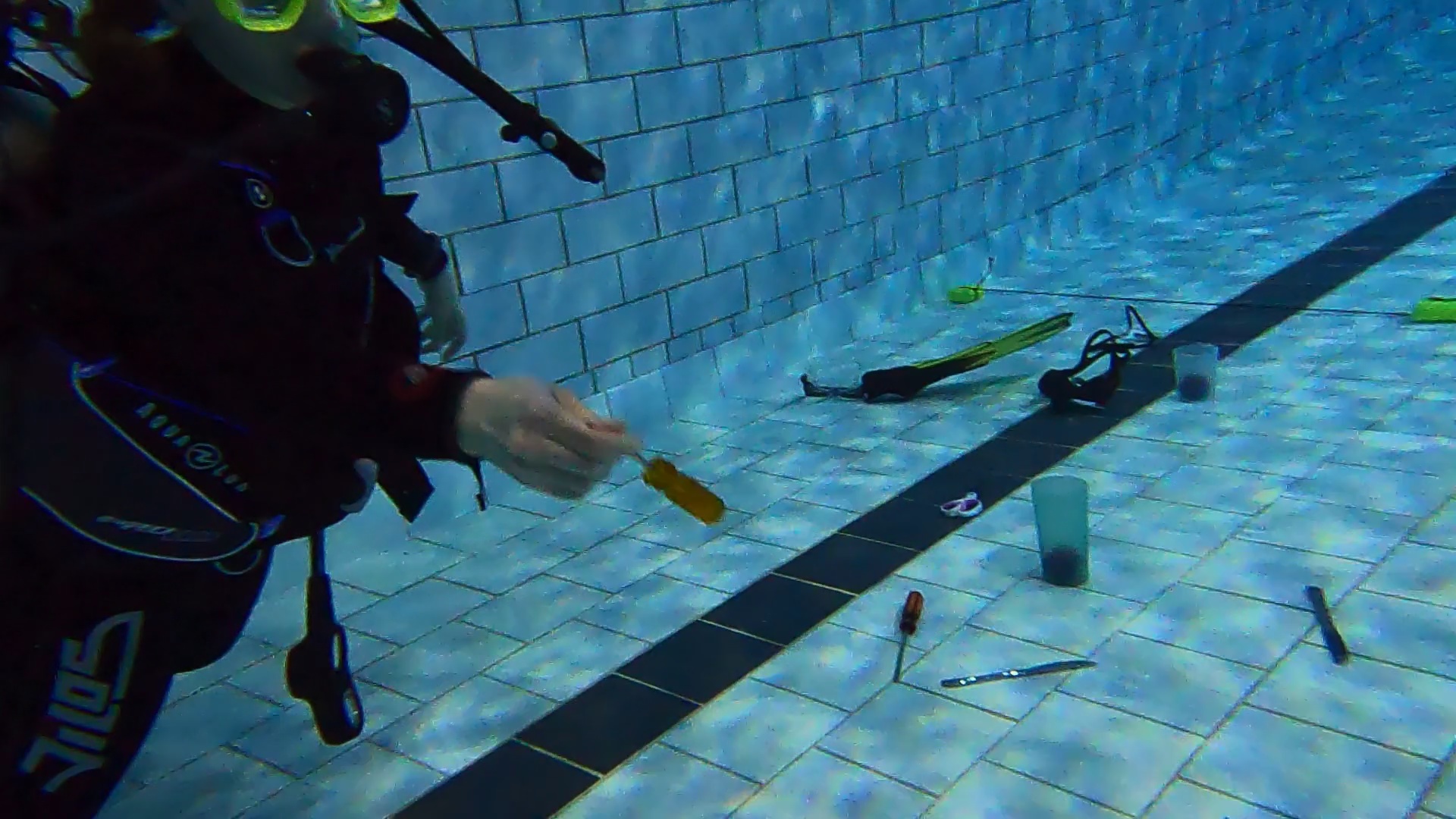}
        \caption{Pool}
        \label{fig:image22}
    \end{subfigure}%
    \begin{subfigure}[b]{0.33\linewidth}
        \centering
        \includegraphics[width=\textwidth]{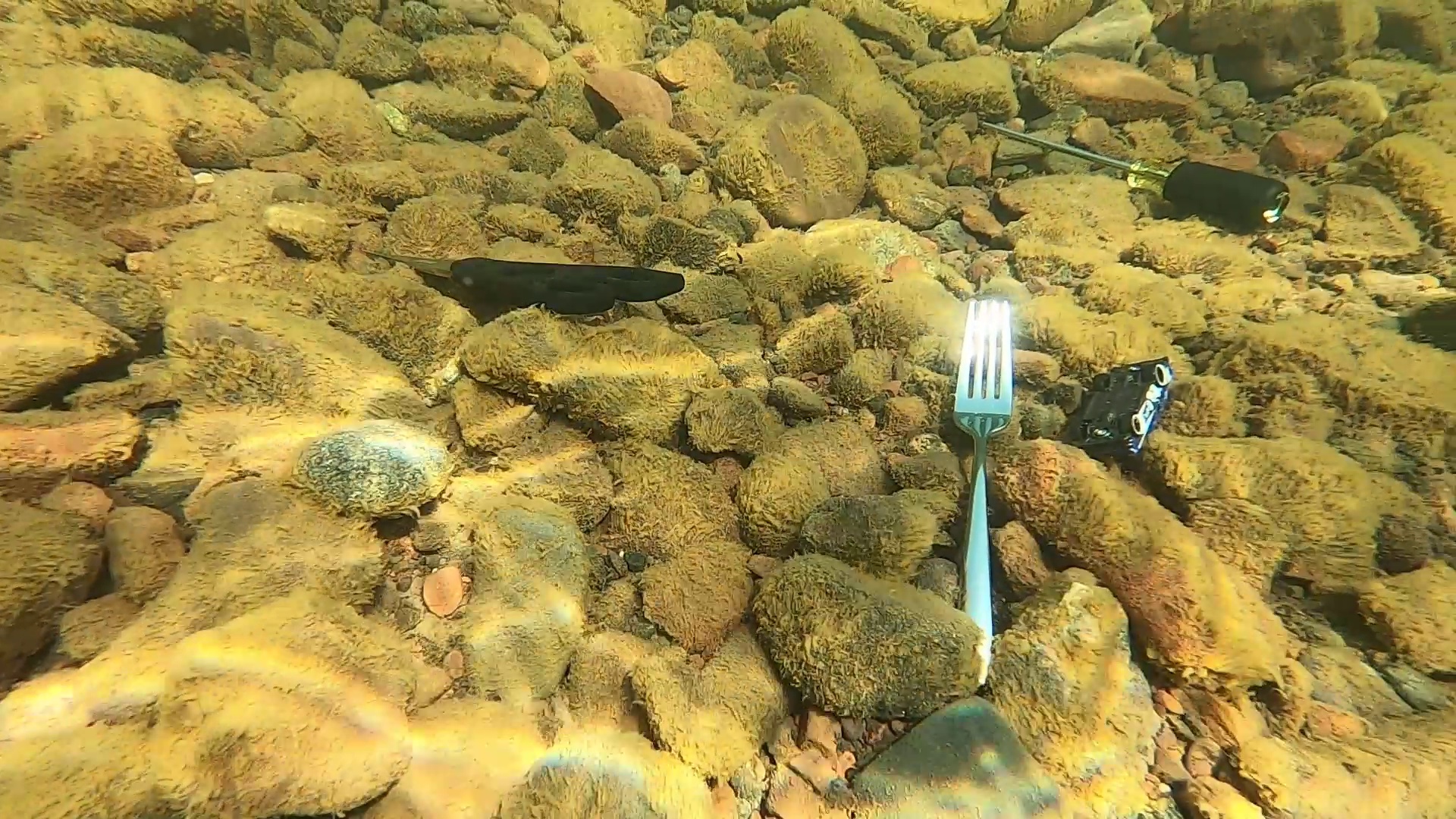}
        \caption{Lake}
        \label{fig:image33}
    \end{subfigure}
    \caption{(Best viewed at 200\% zoom) Examples of images contained in COU across different environments. The variations in lighting, hue, and object distribution make it possible to robustly train deep underwater object detectors with COU's annotated imagery.}
    \label{fig:three-images}
\end{figure}

With the goal of instance segmentation in mind, we focus on getting data across various environments with different lighting conditions. 
Data was collected from three different environments: 
\begin{enumerate*}[label=(\roman*), itemsep=0pt, topsep=0pt] 
\item in a well-lit pool setting, 
\item two lake locations with low visibility underwater, 
\item and in the ocean with mixed-lighting conditions. 
\end{enumerate*}
We captured the pool footage at maximum depth of $4.57$ m , 
with ideal lighting conditions and little to no turbidity in the water (see Fig. \ref{fig:three-images}). 
We collected the open-water images from two Minnesota lakes: Lake Superior located in Duluth, and Green Lake located in Spicer, Minnesota. These lakes are easily accessible for diving and offer visibility that extends beyond a couple of meters.
In these lakes, the lighting conditions are a mix of lowly-lit and moderately-lit conditions, while the water has a high level of turbidity. 
The ocean footage was captured off the coast of Barbados, where the environment has better lighting conditions than the lakes, but not the pool, and the water quality presents very little turbidity. 
The footage was captured using a GoPro\texttrademark{} camera at a resolution of $1920\times 1080$, and $30$ frames per second (FPS), and with a linear lens setting.

\begin{figure}[h]
    \centering
    \includegraphics[width=0.48\textwidth]{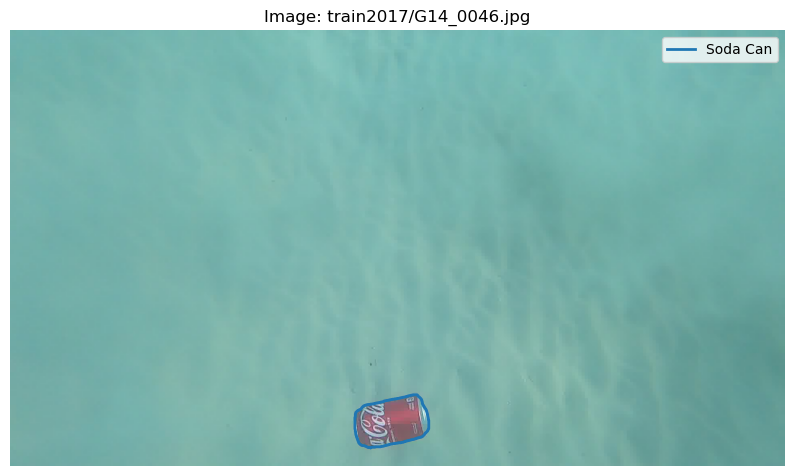}
    \hspace{4pt}
    \includegraphics[width=0.48\textwidth]{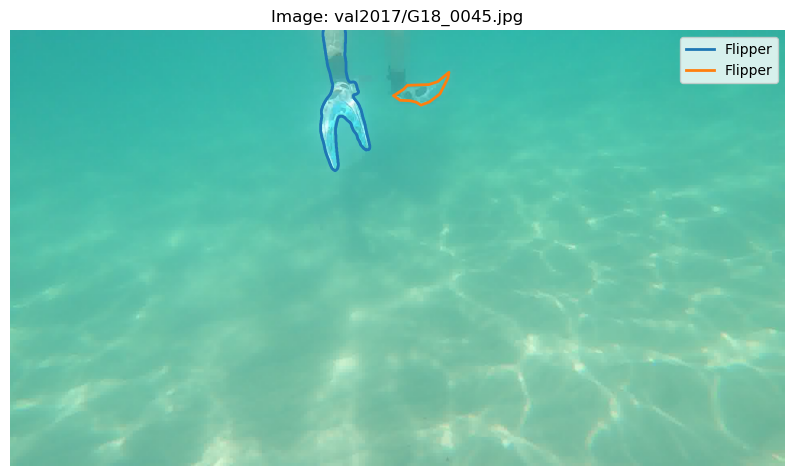}
    \caption{(a) An example of an \textit{isolated} class instance, where the object is present in isolation away from other class instances and other foreground elements. (b) An example of an \textit{in-context} class instance, where the object is present in context, in this case the flippers worn by a swimmer.}
    \label{fig:two-images}
\end{figure}

The footage captured has a mix of \emph{in-context} and \emph{isolated} instances of the classes. An example of an in-context class instance is a flipper on a diver, whereas the isolated variant is the flipper on the ocean floor or an object in isolation like a can. 
Fig. \ref{fig:two-images} illustrates the distinction between the two types, where $11.53\%$ of the images are from the ocean, $38.20\%$ are from the pool, and $50.27\%$ are from lakes.
We collected all these images during underwater robot field trials over a period of $18$ months.

\subsection{Labeling Process}
\begin{figure}[htbp]
    \centering
    \includegraphics[width=0.5\textwidth]{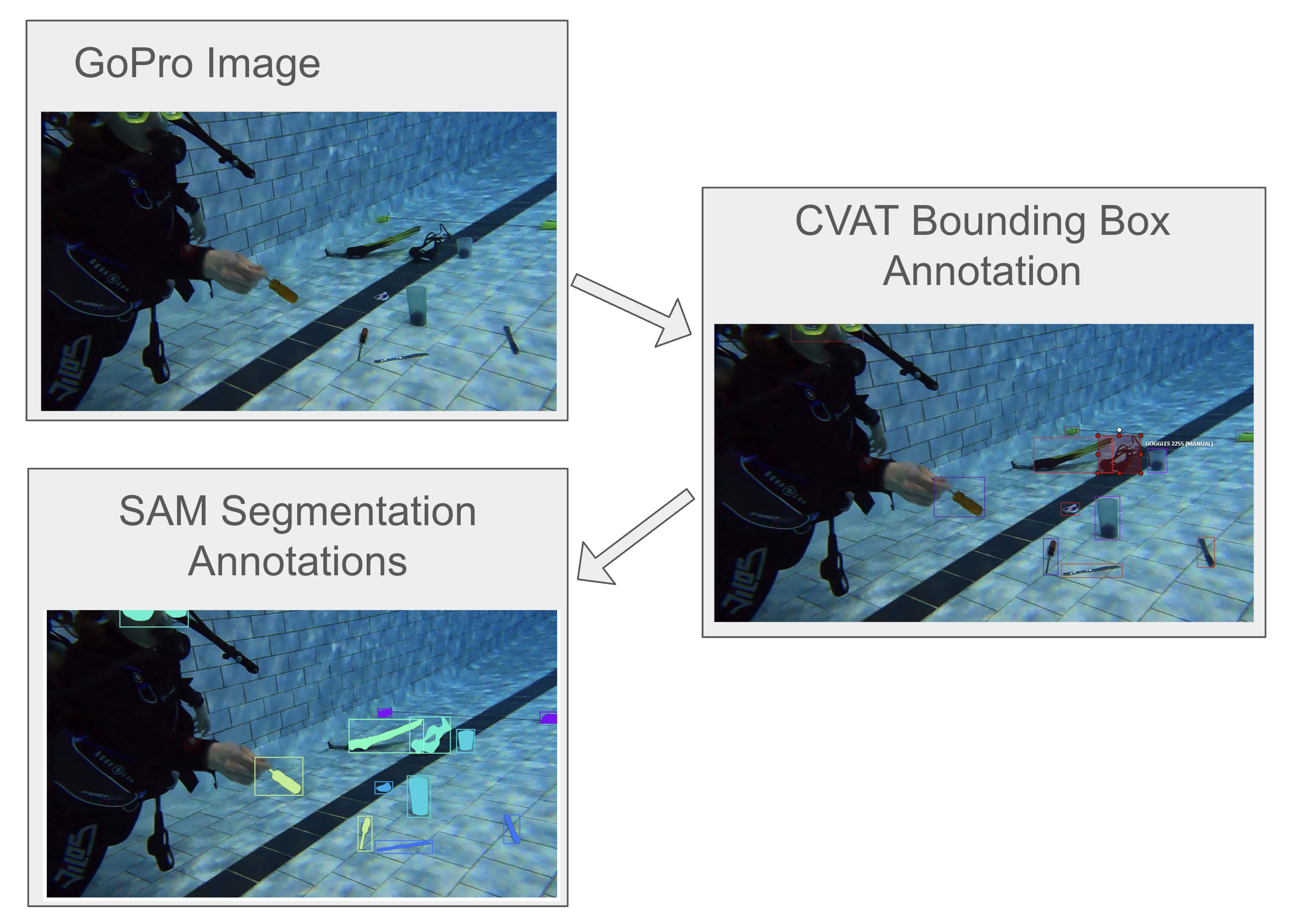}
    \caption{The annotation pipeline used for labeling. GoPro images were manually annotated with bounding boxes using CVAT. These annotated were segmented with SAM.}
    \label{fig:cvat-sam}
\end{figure}

The labeling process of COU is completed in two phases. During the first phase, the images are manually annotated with bounding box annotations using the CVAT labeling tool \cite{cvat}.
Next, we use the Segment Anything Model (SAM) \cite{SAM} to automate the segmentation process by passing bounding box annotations along with the images as illustrated in Fig.~\ref{fig:cvat-sam}.
Note that it takes roughly $12-15$ seconds to annotate a single bounding box, while a segmentation mask can take over $2$ minutes on average. 
To speed up the process, we automate the segmentation process using SAM as it is particularly good at \emph{zero-shot} generalization. This means it can perform segmentation on data it has not been trained on. With the added benefit of passing in the bounding boxes, the images are annotated with a high degree of accuracy. 
The main hardware used was a Nvidia 4080M  GPU, and the processing time for the SAM model was $0.56$ seconds per image. 

\subsection{Post-Processing}
Once the images are segmented, we begin the process of proofreading and evaluation of the annotations. Some of the annotations were mislabeled or had overlapping annotations, which we fixed in CVAT. 
Then we split the dataset into train, test, and validation directories in a $70/20/10$ split, respectively. 
We use the YOLO \cite{wang2024yolov9learningwantlearn} polygon segmentation as our primary annotation format.

\section{Experiments}
\label{sec:models}

To assess the accuracy of models trained on COU and their computational efficiency, We conduct three types of experiments: Benchmarking experiments, Efficiency experiments, and Field robot experiments. Additionally, we perform a direct comparison of models trained on a terrestrial dataset with those on COU, and on a mixture of COU and a terrestrial dataset.

\subsection{Benchmarking Experiments}
We benchmark the dataset on three state-of-the-art deep learning models: YOLOv9 \cite{wang2024yolov9learningwantlearn}, Mask R-CNN \cite{he2018maskrcnn}, and Mask2Former \cite{mask2former}. We pick these models to provide a variety in the architecture used: YOLOv9 uses Generalized Efficient Layer Aggregation Network (GELAN) architecture, Mask-RCNN uses a ResNet-50 backbone, and Mask2Former uses a transformer decoder in its architecture. 
We evaluated two sets of weights per model. 
The first set came from a model trained from scratch. 
For the other set of weights, we fine-tuned a preexisting COCO checkpoint. 
This is a popular technique used for deep object detection underwater that allows leveraging the features extracted from larger datasets. 
\cite{vddc} does this by fine-tuning a model using initial weights from \cite{imagenet}. 
\cite{data_2class} uses a model checkpoint trained on \cite{trash19} as a starting point for fine-tuning. 
We train models using high-performance GPUs for training and evaluation.

\subsubsection{YOLOv9}
YOLOv9 is a single-stage object detector part of the You Only Look Once (YOLO) family of models. 
This iteration introduces Programmable Gradient Information (PGI) to address information loss in deep neural networks. 
This, coupled with their new GELAN architecture, allows YOLOv9 to achieve high accuracy while maintaining efficiency. 
These additions specifically benefit the accuracy and efficiency of the compact models (particularly with issues tracking multiple class instances across frames), which are optimized for OCUs (Onboard Compute Unit). 
We use two models for testing: YOLOv$9$-C and YOLOv$9$-E. 
YOLOv$9$-C is a compact model which can be used on embedded computing platforms while YOLOv$9$-E is an expanded model capable of stronger inference at higher computing cost. 
Models are trained on RTX $4080$M GPU for $12$ hours each. We train the model for 50 epochs with a batch size of 6 and a learning rate of $0.000357$ using the AdamW optimizer.

\subsubsection{Mask R-CNN}
Mask R-CNN generates high-quality segmentation masks for each instance. While it is not suited for real-time robotic applications, we include Mask R-CNN in our evaluation to represent our dataset's upper bound of combined detection and segmentation accuracy. 
We implement Mask R-CNN using Detectron2 \cite{wu2019detectron2}, with a ResNet-$50$-FPN backbone. The ResNet-$50$-FPN backbone is a mid-size variant of the Residual Network \cite{ResNet} family of models that provides depth in terms of accuracy while maintaining efficiency. We choose this backbone instead of ResNet-$101$ or ResNet-$32$ and below to balance accuracy with efficiency. Mask R-CNN adds only a small overhead to its object detection counterpart Faster R-CNN \cite{faster-rcnn}, making it an attractive option for scenarios where precise object boundaries are crucial. 
The batch size and learning rate are tuned to be $128$ and $0.00025$ respectively. We train the model for 50 epochs on an L$4$ GPU using Google Colab \cite{google_colab}. 

\subsubsection{Mask2Former}
Masked-attention Mask Transformer (Mask2Former) is a deep learning model designed for performing any image segmentation task (panoptic, instance, or semantic). 
It uses ResNet-$50$ backbone for feature extraction, a pixel decoder, and a transformer decoder. 
Instead of cross-attention, it incorporates masked attention which leads to faster convergence and improved performance compared to standard transformer decoders. 
We train the model on a RTX 2080 GPU for
$50$ epochs, with batch size of $8$ and a base learning rate of $1.0e-05$.

\begin{table*}[t]
    \centering
    \begin{tabular}{|c|c|c|c|c|c|}
    \toprule
         \textbf{Dataset} & \textbf{Images} & \textbf{Total classes} & \textbf{Class type} & \textbf{Annotation} & \textbf{Environment}\\ \hline \hline
         COU & \textbf{9,757} & 24 & Object & \makecell {\ Bounding box \\ Segmentation} & Ocean, Lake, Pool\\ \midrule
         TrashCan & 7,212 & 22 & Object, ROV, Marine life & \makecell {\ Bounding box \\ Segmentation} & Ocean\\
         DeepPlastic & 3,174 & 1 & Object & Bounding box & Ocean, Lake\\ \midrule
         Seaclear & 8,610 & 40 & Object, ROV, Marine life & \makecell {\ Bounding box \\ Segmentation} & Ocean\\ \midrule
         UNO & 5,902 & 22 & Object, ROV, Marine life & \makecell {\ Bounding box \\ Segmentation} & Ocean\\
         \bottomrule
    \end{tabular}
    \caption{Comparison of existing underwater object datasets with COU. COU contains a variety of environments and contains the highest number of images among the datasets.}
    \label{tab:data_comp}
\end{table*}

\subsection{Efficiency Experiments}
We measure the efficiency of each model based on two factors: inference speed and hardware used. The hardware used is important for practical reasons; many ROVs in the field are designed with compact OCUs that lack the compute power required for large-scale models. Thus it is essential to measure the performance of the models used on comparable hardware to that of the field. We test the networks on two devices, a Nvidia Jetson Orin `NX', and a Nvidia RTX 2080 Ti. The Jetson is a popular computing platform for robotic systems. 
When we say semi real-time segmentation, we are loosely referring to 5 FPS and higher. 
While no platform-specific optimizations are applied for the YOLO models, \cite{wang2024yolov9learningwantlearn} provides some parameters that can compress the model further; those guidelines are followed. Additionally the Tao Toolkit by Nvidia \cite{tao_toolkit} is used to run the detectron models on the Jetson Orin \cite{mask2former}\cite{he2018maskrcnn}. 

\subsection{Field Robot Experiments}
\begin{figure}[h]
    \centering
    \includegraphics[width=0.95\columnwidth]{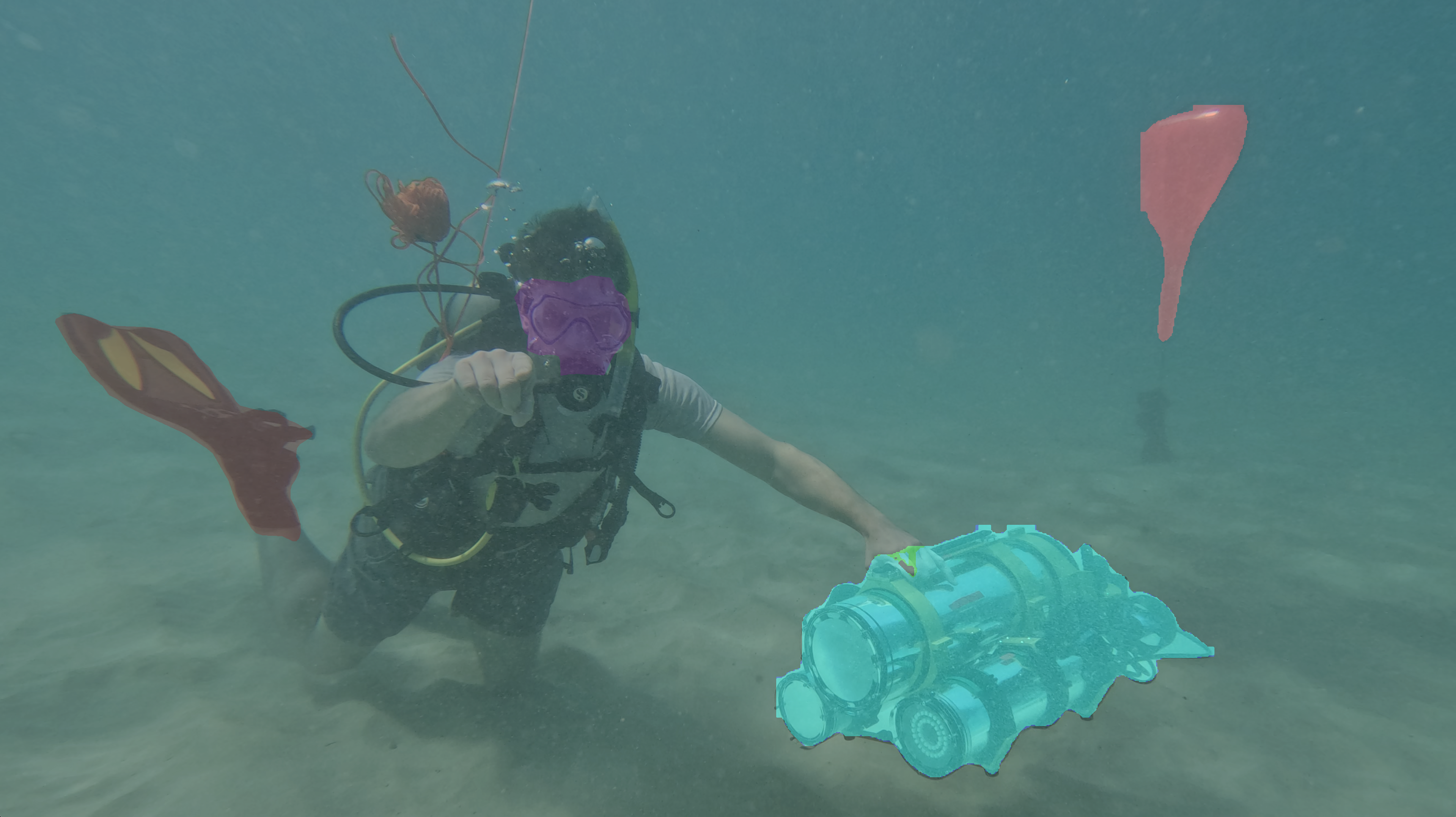}
    \caption{Predicted image using a YOLOv9c model in the ocean environment. The model is pretrained on COCO and fine-tuned using COU. The object segmentations in the frame (left to right) are for Flipper, Goggles, and LoCO class instances. The object on the right is a dive bag that is being identified as a flipper.}
    \label{fig:cou-seg}
\end{figure}

To assess the real-world applicability of the dataset, we conduct qualitative evaluations using unseen data from ocean environments (Fig.~\ref{fig:cou-seg}).
For this, we use YOLOv9-C model deployed on the MeCO AUV \cite{meco}. 
YOLOv9-C is trained on a fine-tuned COU dataset with COCO classes using transfer learning \cite{TL}. 
The convolution layers are frozen and only the classification layers are retrained with a subset of COCO. 
The classes are remapped to match the COCO classes where there is an overlap and the rest of the classes were appended to the class list. 
This is done to preserve the classes in COCO that are relevant to the environment along with additional classes from COU.

\subsection{Terrestrial data comparison}
We use YOLOv9-C to compare the performance of the COCO dataset with COU for underwater imagery. We use three different training approaches to compare the performance of terrestrial models to underwater models. The \textit{first} model is an off-the-shelf model pre-trained only on COCO. This represents a terrestrial model. 
The \textit{second} model fine-tunes the terrestrial model by training only the classification layers with COU. 
We preserve the classes by matching overlapping class IDs and using unique class IDs for non-overlapping classes. 
The \textit{third} model follows the traditional fine-tuning process by training the entire architecture with COU. We test all three models on the COU Test split.

\section{Results}
\label{sec:results}
\subsection{Dataset Comparisons}

We compare the COU dataset with existing underwater datasets containing similar objects. 
We exclude terrestrial object datasets from this comparison, since the environment conditions in the underwater domain limit the feasibility of gathering data at the same scale of terrestrial datasets. 
Moreover, we do not include underwater datasets containing only marine life species, as the objective of COU is to encompass marine litter and tools used by divers. 

Table \ref{tab:data_comp} shows that COU is the largest underwater object dataset containing both bounding box and segmentation labels. Furthermore, it exhibits a great variety in the environments when compared to other datasets. The only dataset containing more classes than COU is Seaclear which includes 40 classes. However, it is important to note that these classes cover marine life, debris, and ROV parts, which differ from the scope of the COU dataset.

\begin{table*}[htbp]
    \centering
    \small
    \setlength{\tabcolsep}{3pt}
    \begin{tabular}{lccc|ccc|ccc|ccc}
        \multirow{3}{*}{} & \multicolumn{6}{c|}{Tested on COU: val} & \multicolumn{6}{c}{\textbf{Tested on COU: test}} \\
         & \multicolumn{3}{c|}{Trained from Scratch} & \multicolumn{3}{c|}{Finetuned COCO model} & \multicolumn{3}{c|}{Trained from Scratch} & \multicolumn{3}{c}{Finetuned COCO model} \\
         \textbf{Model} & \textbf{AP}$_{50}$ & \textbf{AP}$_{75}$ & \textbf{mAP$_{.5-.95}$} & \textbf{AP}$_{50}$ &  \textbf{AP}$_{75}$ & \textbf{mAP$_{.5-.95}$} & \textbf{AP}$_{50}$ &  \textbf{AP}$_{75}$ & \textbf{mAP$_{.5-.95}$} & \textbf{AP}$_{50}$ &  \textbf{AP}$_{75}$ & \textbf{mAP$_{.5-.95}$} \\
        \hline 
         & \multicolumn{3}{c|}{} & \multicolumn{3}{c|}{} & \multicolumn{3}{c|}{} & \multicolumn{3}{c}{} \\
        \textbf{YOLOv9-C} & 0.754 & 0.638 & 0.595 & 0.788 & 0.699 & 0.644 & 0.827 & 0.689 & 0.645 & 0.828 & 0.694 & 0.646 \\
        \textbf{YOLOv9-E} & 0.738 & 0.643 & 0.609 & 0.845 & 0.743 & 0.725 & 0.754 & 0.640 & 0.604 & 0.826 & 0.705 & 0.706 \\
        \textbf{Mask R-CNN} & 0.802 & 0.751  & \textbf{0.713} & 0.818 & 0.763 & 0.720 & 0.811 & 0.769 & 0.702 & 0.849 & 0.810 & 0.773 \\
        \textbf{Mask2Former} & 0.817 & 0.755 & 0.705 & 0.841 & 0.786 & \textbf{0.732} & 0.850 & 0.807 & \textbf{0.728} & 0.872 & 0.839 & \textbf{0.796} \\
        \hline
    \end{tabular}
    \caption{Comparison of model performance on the validation and testing set of COU using mask AP scores. mAP$_{.5-.95}$ denotes the average mAP score.}
    \label{tab:model-comparison}
\end{table*}

\subsection{Accuracy}
To evaluate the accuracy of the models, we find the Average Precision (AP) of the models on pixel-wise segmentation. The models output a confidence score for each instance identified (\emph{recall value/threshold}). The AP is calculated by taking the mean of precision values, each corresponding to a different recall threshold. These precision values are weighted based on the incremental gain in recall from one threshold to the next. The AP at a specific threshold is important because these threshold values can be adjusted during inference time.

Table \ref{tab:model-comparison} compares the benchmarking performance across the models on COU's validation and test sets. 
The left two columns contain results on the validation set. 
The Mask R-CNN model has the highest mAP@.5-.95 in the models trained from scratch, while the Mask2Former model has the highest accuracy in the fine-tuned (COCO model fine-tuned on COU) category. 
In the test set, the Mask2Former model maintains the highest accuracy both when trained from scratch and when fine-tuning a pre-trained COCO model as a base. 
Noticeably, the YOLO models perform worse on average than the other two architectures, trading off performance for accuracy, compared to Mask R-CNN and Mask2Former which are known for precise segmentations.

\begin{table}[htbp]
    \centering
    \begin{tabular}{lc}
         \textbf{COCO Model}  & \textbf{mAP$_{\text{\textbf{.5-.95}}}$} \\ \midrule
         YOLOv9-C  &  0.030\\
         YOLOv9-C-COU-class  & 0.583 \\
         YOLOv9-C-COU-conv  & 0.646 \\
    \end{tabular}
    \caption{Comparison of three pretrained YOLOv9-C COCO models on COU Test(top to bottom): the first model is only trained on COCO, the second model has the classification layers fine tuned with COU, the third model fine tunes the convolution layers with COU.}
    \label{tab:COCO-results}
\end{table}

Table \ref{tab:COCO-results} compares the accuracy of three training methods for the YOLOv9-C model.
From our results on the COU test set, we find that the terrestrial model performs significantly worse than the latter two models. 
The YOLOv9-C-COU-class (only trained the classification layers) has much better accuracy than the first model but performs worse than YOLOv9-C-COU-conv. The YOLOv9-C-COU-conv model shows the highest accuracy out of all the models. The difference between the two fine-tuned models can be explained by the nature of the images in the underwater domain. Images underwater tend to be more blurry, suffer from water turbidity, and exhibit varying lighting conditions making the objects appear less distinctly in the image. 

\subsection{Efficiency}

\begin{table}[htbp]
    \centering
    \begin{tabular}{lcc}
         \textbf{Model} & \textbf{Jetson Orin Nx} & \textbf{RTX 2080 Ti} \\ \midrule
         YOLOv9-C & 150 ms/img &  34 ms/img\\
         YOLOv9-E & 331 ms/img & 45 ms/img \\
         Mask R-CNN & 390 ms/img & 60 ms/img \\
         Mask2Former & 469 ms/img & 131 ms/img\\
    \end{tabular}
    \caption{Inference time benchmark on Jetson Orin NX and RTX 2080 Ti}
    \label{tab:results}
\end{table}

In Table \ref{tab:results}, we see the inference times across the two platforms. 
The compact YOLOv9-C model shows the fastest response time on both hardware platforms. Crucially, it has an inference time of $150$ milliseconds on the Jetson Orin, which equates to roughly $6-7$ FPS.
This satisfies the semi-real time criterion. 
The larger YOLOv9-E model halves the inference speed ($331$ milliseconds, or $3$ FPS) on the Jetson Orin making it not suitable for semi real-time deployment. 
The larger detectron \cite{wu2019detectron2} models: Mask-RCNN, and Mask2Former, have considerably slower inference times on both platforms. 
Mask-RCNN runs at $390$ milliseconds per image, or $2.7$ FPS, while Mask2Former runs at $469$ milliseconds per image or $2.3$ FPS. 
While we are able to run both YOLO models on the Jetson without platform-specific optimizations, running the two detectron models required some optimizations. 
The recommended VRAM for either of the models is $12$ gigabytes, while the Jetson use has $8$ gigabytes of VRAM. 
Consequently, we use the Tao toolkit by Nvidia \cite{tao_toolkit} to run the inference tests on the Jetson.

\section{Conclusion}
In this paper, we present COU: a dataset of common objects underwater. In comparison to commonly used pre-existing datasets such as \cite{trash19}, COU brings a $61\%$ increase in total images, as well as a $60\%$ increase in the number of classes. The dataset also contains high-resolution images. With the improvements of SOTA models like \cite{wang2024yolov9learningwantlearn}, particularly in terms of efficiency, higher resolution images can be inferenced in real-time, on edge computing devices/onboard compute units. The accuracy of these models has been evaluated and provides a benchmark for other models on the dataset.

\bibliographystyle{IEEEtran}
\bibliography{ref}

\end{document}